\documentclass[11pt,a4paper]{article}
\usepackage[hyperref]{acl2021}
\usepackage{times}
\usepackage{latexsym}

\usepackage{microtype}

\aclfinalcopy 

\usepackage{helvet} 
\usepackage{courier}  
\usepackage{graphicx} 
\usepackage{amsmath}
\usepackage{multirow}
\usepackage{lettrine}
\usepackage[switch]{lineno}
\usepackage{natbib}  
\usepackage{caption} 
\usepackage{makecell}
\usepackage[normalem]{ulem}
\usepackage{color}
\usepackage{cleveref}
\usepackage{mathrsfs}

\usepackage{latexsym}
\usepackage{amsfonts}
\usepackage{stmaryrd}
\usepackage{subcaption}
\usepackage{booktabs}
\usepackage{url}
\usepackage{bm}

\title{Decoupled Dialogue Modeling and Semantic Parsing for Multi-Turn Text-to-SQL}

\author{Zhi Chen, Lu Chen\thanks{The corresponding authors are Lu Chen and Kai Yu.}, Hanqi Li, Ruisheng Cao, Da Ma,
 \textbf{Mengyue Wu} \and \textbf{Kai Yu}\footnotemark[1] \\
        X-LANCE Lab, Department of Computer Science and Engineering \\ 
        MoE Key Lab of Artificial Intelligence, AI Institute, Shanghai Jiao Tong University\\
        Shanghai Jiao Tong University, Shanghai, China\\
    State Key Lab of Media Convergence Production Technology and Systems, Beijing, China\\
    \texttt{\{zhenchi713, chenlusz, kai.yu\}@sjtu.edu.cn}
        }

\date{}

\begin{document}
\maketitle
\begin{abstract}
Recently, Text-to-SQL for multi-turn dialogue has attracted great interest. Here, the user input of the current turn is parsed into the corresponding SQL query of the appropriate database, given all previous dialogue history. Current approaches mostly employ end-to-end models and consequently face two challenges. First, dialogue history modeling and Text-to-SQL parsing are implicitly combined, hence it is hard to carry out interpretable analysis and obtain targeted improvement. Second, SQL annotation of multi-turn dialogue is very expensive, leading to training data sparsity. In this paper, we propose a novel decoupled multi-turn Text-to-SQL framework, where an utterance rewrite model first explicitly solves completion of dialogue context, and then a single-turn Text-to-SQL parser follows. A dual learning approach is also proposed for the utterance rewrite model to address the data sparsity problem. Compared with end-to-end approaches, the proposed decoupled method can achieve excellent performance without any annotated in-domain data. With just a few annotated rewrite cases, the decoupled method outperforms the released state-of-the-art end-to-end models on both SParC and CoSQL datasets.
\end{abstract}

\section{Introduction}
\begin{figure}[t]
\centering
\includegraphics[width=0.5\textwidth]{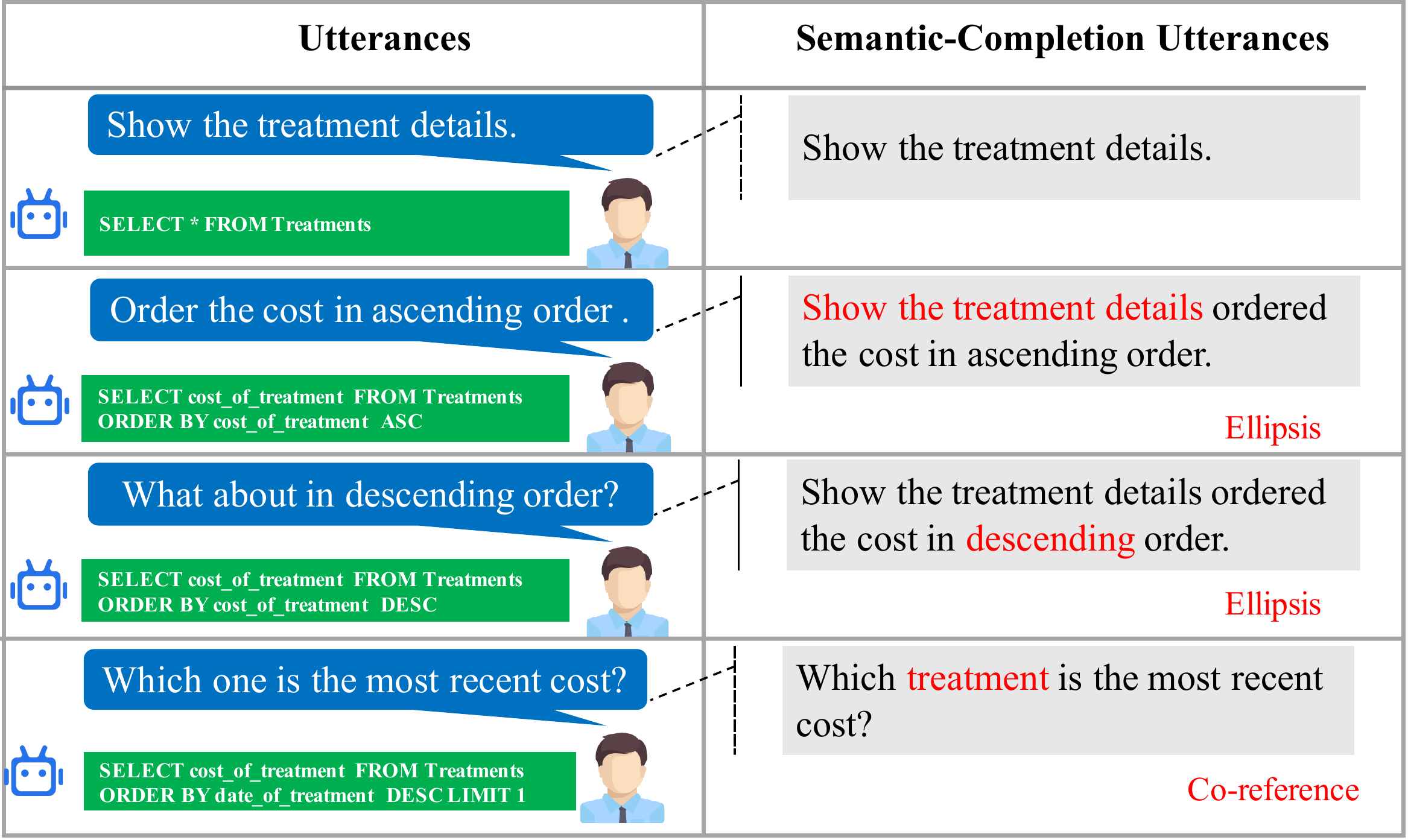} 
\caption{An example to demonstrate the co-reference and ellipsis phenomenon in a conversation, where the right column shows the annotated semantic-completion utterances.}
\label{fig:example}
\end{figure}

Text-to-SQL has lately become an interesting research topic along with the high demand to query a database using natural language (NL). 
Standard large database format can only be accessed with Structured Query Language (SQL), which requires certain special knowledge from users, hence lowering the accessibility of these databases. 
Text-to-SQL tasks, however, greatly minimize this gap and allow the query based on NL. 
Previous work on Text-to-SQL mostly focuses on single-turn utterance inference, evaluated on context-independent Text-to-SQL benchmarks. 
Nevertheless, in practice, the users usually need to interact with the Text-to-SQL system step-by-step to address their query purpose clearly. 
Under such conversation scenarios, the co-reference and information ellipses are always present, shown in Figure~\ref{fig:example}. Recently proposed methods are mostly end-to-end, which endeavors to design a suitable model to encode the dialogue context and infer the corresponding SQL based on the whole dialogue context. 
The main limitation of the end-to-end multi-turn Text-to-SQL models lies in their extreme reliance on annotated multi-turn Text-to-SQL data. The large-scale multi-turn Text-to-SQL data is time-consuming and expensive. 
The annotators not only need to be SQL experts but also have to infer the complete and exact query intent of the latest utterance of the speaker.

Different from previous end-to-end approaches, we propose a \textbf{DE}coupled mu\textbf{L}ti-\textbf{T}urn p\textbf{A}rsing (\textbf{DELTA}) framework, which decouples the multi-turn Text-to-SQL into two subsequent pipeline tasks: utterance rewrite and single-turn Text-to-SQL. In recent years, these two individual tasks are both well-studied. The utterance rewrite task aims to generate the latest semantic-completion question based on the dialogue context. The single-turn Text-to-SQL task aims to parse the semantic-completion question to a SQL, where the state-of-the-art methods~\cite{shi2020learning,yu2020grappa,chen-etal-2021-shadowgnn,rubin-berant-2021-smbop} can achieve over 70\% exact match accuracy on Spider~\cite{yu2018spider} (a cross-domain single-turn Text-to-SQL dataset) and even achieve more than 80\% on easier Text-to-SQL benchmarks~\cite{dahl1994expanding, zhongSeq2SQL2017}. 
However, there is no rewrite data on the existing multi-turn Text-to-SQL benchmarks and the existing utterance rewrite datasets normally pay more attention to the co-reference problem but ignore the information ellipses. 
Due to the limitation of the in-domain annotated rewrite data, we further propose a dual learning method to make comprehensive use of the unlabeled multi-turn data to learn a reliable rewrite model. 
Our proposed framework DELTA is evaluated on both the SParC~\cite{Yu&al.19} and CoSQL datasets~\cite{yu2019cosql}, the two existing large-scale benchmark for the multi-turn Text-to-SQL task.

Contributions are highlighted below:

\begin{itemize}
\item We propose a decoupled parsing framework for the multi-turn Text-to-SQL task, whose annotated data is much easier to collect.
Even without any in-domain multi-turn Text-to-SQL data, the decoupled parsing method can achieve encouraging results on multi-turn Text-to-SQL benchmarks.
\item The decoupled framework includes an utterance rewrite model which is adapted from the pretrained BART~\cite{lewis2020bart}, with a newly implemented dual learning method to make comprehensive use of the unlabeled multi-turn data. 
Our adapted rewrite model achieves new state-of-the-art performance on the utterance rewrite benchmarks.
\item With fully labeled multi-turn Text-to-SQL data, our decouple parsing method outperforms all the released end-to-end multi-turn Text-to-SQL model.
\end{itemize}




\section{Decoupled Parsing Framework}
In this section, we elaborate our decoupled parsing framework, which consists of two phases: 1) an \textit{utterance rewrite model} (\Cref{ssec:phase1}), to generate semantic-completion question based on the dialogue context; 2) a \textit{single-turn Text-to-SQL parser} (\Cref{ssec:phase2}), which is fed with the rewritten question to predict the corresponding SQL query.
To further improve the rewrite model performance, we propose a dual learning method to make use of large-scale unlabeled data, which is detailed in \Cref{sec:dual}. 

\subsection{Phase-\uppercase\expandafter{\romannumeral1}: BART as Rewrite Model}
\label{ssec:phase1}
We leverage the pretrained BART~\cite{lewis2020bart}, which is a Transformer-based encoder-decoder architecture, as the utterance rewrite model. This idea is inspired by its success on the text generation tasks, including question answering and summarization. 
Along with the success of pretrained language models, the Transformer architecture has been widely applied in natural language process (NLP) tasks. Transformer aims to encode a sequence $X=[x_i]_{i=1}^{n}$ with the \textit{self-attention} mechanism~\cite{vaswani2017attention}. Assume that $[\mathbf{x}_i^{(l)}]_{i=1}^{n}$ is the representation of the sequence $X$ at $(l)$-th Transformer layer. The next Transformer takes the following operations with $H$ attention heads:
\begin{gather}
\alpha_{ij}^{(h)} = \mathop{\rm softmax}\limits_{j}\left(\frac{\mathbf{x}_i^{(l)}\mathbf{W}_Q^{(h)}\left(\mathbf{x}_j^{(l)}\mathbf{W}_K^{(h)}\right)^{\top}}{\sqrt{d_z/H}}\right), \notag \\
\mathbf{z}_i = \bigparallel_{h=1}^{H}\sum_{j=1}^n \alpha_{ij}^{(h)} \mathbf{x}_j^{(l)}\mathbf{W}_V^{(h)}, \notag  \\
\bar{\mathbf{x}}_i^{(l+1)} = \mathop{\rm LN}\left(\mathbf{x}_i^{(l)} + \mathbf{z}_i\right),  \notag \\
\mathbf{x}_i^{(l+1)} = \mathop{\rm LN}\left(\bar{\mathbf{x}}_i^{(l+1)} + \mathop{\rm FFN}\left(\bar{\mathbf{x}}_i^{(l+1)} \right)\right),
\label{eq:self_attention1}
\end{gather}
where $h$ is the head index, $d_z$ is the hidden dimension, $\alpha_{ij}^{(h)}$ is attention probability, $\bigparallel$ denotes the concatenation operation, $\mathop{\rm LN}(\cdot)$ is layer normalization~\cite{ba2016layer} and $\mathop{\rm FFN}(\cdot)$ is a feed-forward network consists of two linear transformations.

Similar to other large-scale pretrained language models, BART also uses a standard Transformer-based sequence-to-sequence architecture, where the encoder is the bidirectional Transformer and the decoder is the auto-regressive Transformer. BART's pretraining method reconstructs the original text from its corrupted text. In its essence, BART is a denoising autoencoder, which is applicable to a very wide range of NLP tasks. In our utterance rewrite task, both the co-reference and information ellipses can be regarded as the corrupted noise of an utterance. Based on this idea, BART can be an appropriate method to denoise the co-reference and information ellipses. 
 In addition, the rewrite data in the public multi-turn Text-to-SQL benchmarks are lacking. 
Therefore, we propose a dual learning method to learn a reliable rewrite model with large-scale unlabeled dialogue data. The details are introduced in Section~\ref{sec:dual}.

\subsection{Phase-\uppercase\expandafter{\romannumeral2}: RATSQL as Parsing Model}
\label{ssec:phase2}
Given a natural language question and a schema for a relational database, the goal of Text-to-SQL parser is to generate the corresponding SQL query.
Regarding the single-turn Text-to-SQL parsing model, we directly use the current state-of-the-art RATSQL model~\cite{rat-sql}. 
RATSQL provides a unified framework, which is based on a relation-aware Transformer (RAT), to encode the question and the corresponding schema. 
The relation-aware Transformer is an important extension to the traditional transformer, which takes the input sequence as a labeled, directed, fully-connected graph. The pairwise relations between input elements are considered in RAT. RAT incorporates the relation information in Equation~\ref{eq:self_attention1}. The edge from element $\mathbf{x}_i$ to element $\mathbf{x}_j$ is represented by vector $\mathbf{r}_{ij}$, which is represented as biases incorporated in the Transformer layer, as follows:
\begin{gather*}
\alpha_{ij}^{(h)} = \mathop{\rm softmax}\limits_{j}\left(\frac{\mathbf{x}_i^{(l)}\mathbf{W}_Q^{(h)}\left(\mathbf{x}_j^{(l)}\mathbf{W}_K^{(h)}+\mathbf{r}_{ij}\right)^{\top}}{\sqrt{d_z/H}}\right), \notag \\
\mathbf{z}_i = \bigparallel_{h=1}^{H}\sum_{j=1}^n \alpha_{ij}^{(h)} \left(\mathbf{x}_j^{(l)}\mathbf{W}_V^{(h)}+\mathbf{r}_{ij}\right). \notag
\end{gather*}
The relations among the Text-to-SQL input elements can be categorized into three types: intra-question, question-schema, and intra-schema. The intra-question relation means both tokens are the elements of the question. The question-schema relations are normally named by schema linking, which is used to represent the matching degree between the question token and the schema token. 
The intra-schema relations include the relation types of the relational database: primary key, foreign key, etc. However, these relations within the input elements are independent of the domain information of the database. 
Incorporating the domain-independent relations into the representation of the Text-to-SQL input is thus beneficial to the Text-to-SQL parser generation.

During decoding, the SQL query is first represented as an abstract syntax tree (AST) following a well-designed grammar. Followed by that, the AST is flattened as a sequence by the deep-first search (DFS) method. RATSQL uses the LSTM to generate the flattened AST sequence. The generated actions defined by the grammar has two structures: (1) it expands the last generated node into a grammar rule, called A{\small PPLY}R{\small ULE} or when completing a leaf node; (2) alternatively, it selects a column/table from the schema, called S{\small ELECT}C{\small OLUMN} and S{\small ELECT}T{\small ABLE}.

\section{Dual Learning for Utterance Rewrite}
\label{sec:dual}

\begin{figure*}[t]
\centering
\includegraphics[width=\textwidth]{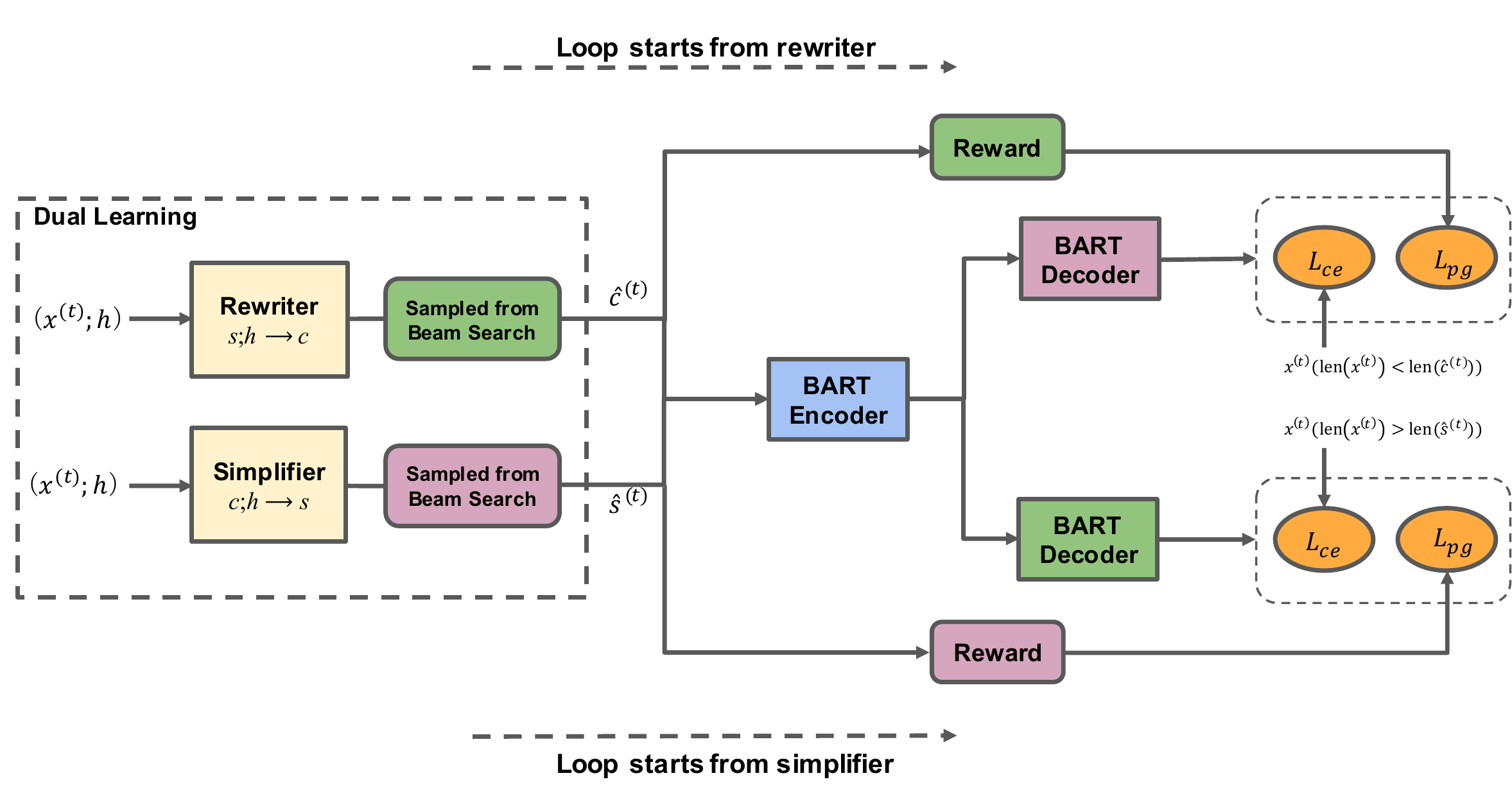} 
\caption{The dual learning architecture of the utterance rewrite task. Rewriter and Simplifier represent the dual models: rewrite model and simplification model. Both two models are initiated by pretrained BART. During the dual learning closed-loop game, two models are updated by policy gradient loss ($L_{pg}$) and cross-entropy loss ($L_{ce}$).}
\label{fig:dual_framework}
\end{figure*}

Due to the limitation of the in-domain annotated rewrite data, we propose a semi-supervised learning method via dual learning to make full use of the unlabeled multi-turn data to learn a reliable rewrite model.
In this section, we first introduce the primal and dual tasks of the utterance rewrite. We then demonstrate the dual learning algorithm for utterance rewrite in detail, where a large amount of unlabeled utterance rewrite data participate in optimizing the primal and dual models under the dual learning framework.

\subsection{Primal and Dual Tasks}
 In a conversation scenario, the co-reference and information ellipses are always present in the user's expressions~\cite{androutsopoulos1995natural}. Recently, \citet{ijcai2020Liu} make a significant step to analyze the co-reference and ellipsis phenomenon~\cite{iyyer2017search,andreas2020task} at the fine-grained level. Co-reference has been divided into five types according to the existing pronoun: \textit{Bridging Anaphora}, \textit{Definite Noun Phrases}, \textit{One Anaphora}, \textit{Demonstrative Pronoun}, and \textit{Possessive Determiner}. Ellipsis has been characterized by its intention: \textit{Continuation} and \textit{Substitution}. where the substitution can be further classified into 4 types: \textit{explicit} vs. \textit{implicit} and \textit{schema} vs. \textit{operator}. The detailed introduction of these fine-grained types refers to \cite{ijcai2020Liu}.
 
 The primal task aims to denoise the above co-reference and ellipsis and generate a semantic-completion utterance $c^{(t)}=[c^{(t,i)}]_{i=1}^n$ based on the utterance $x^{(t)}=[x^{(t,i)}]_{i=1}^m$ at the $t$-th turn and the dialogue history $h=[x^{(j)}]_{j=1}^{t-1}$. We directly use the pretrained BART as the rewrite model (named \textbf{rewriter}). We further concatenate the dialogue history and the latest utterance as the input of the rewriter, where they are separated by the special token ``${\rm \textless /s \textgreater}$". The dual task is to generate a simplified expression based on the latest utterance and the dialogue history. The simplified expression contains the above co-reference and ellipsis as more as possible without changing the original semantic meaning of the dialogue. Similar to the rewriter, we use the pretrained BART as the initial simplification model (named \textbf{simplifier}).

\subsection{Dual Learning Algorithm}
Under the dual learning framework, the dual models can be regarded as two agents in a closed-loop game. The game starts with one of the dual agents. The output of the start agent will be scored by an external reward function. Since the reward feedback is non-differentiable, the start agent is optimized by the policy gradient method~\cite{sutton1999policy}. The end agent is fed with the output of the start agent, where the end agent aims to reconstruct the initial input of the start agent. Thus, the end agent can be optimized by maximum likelihood estimation (MLE). Before deep-diving into the dual learning algorithm, we first introduce the definitions of the dual framework for the utterance rewrite.

\subsubsection{Definition}
Suppose we have unlabeled dialogue data $\mathcal{D}_{u}={\{ (x^{(t)};h)\}}$.
There are two dual models: rewriter with parameter $\Theta_c$ and simplifier with parameter $\Theta_s$. Two language models (${\rm LM}_c(\cdot)$ and ${\rm LM}_s(\cdot)$) are used to evaluate the quality of the generated utterances by rewriter and simplifier respectively. Both two language models are fine-tuned from GPT-2 model~\cite{radford2019language}. ${\rm LM}_c(\cdot)$ is trained with semantic-completion Spider dataset. ${\rm LM}_s(\cdot)$ is trained from multi-turn Text-to-SQL data (SParC and CoSQL), where the utterances at the first turn are removed. There is an external single-turn Text-to-SQL parser ${\rm RATSQL(\cdot)}$, which parses a question into a SQL query. Next, we will introduce the strategy of agent optimization under dual learning framework.

\subsubsection{Loop Starts from Rewriter}
\label{sec:loop1}
As shown in Fig.~\ref{fig:dual_framework}, we sample an unlabeled dialogue data $(x^{(t)};h)$ from $\mathcal{D}_{u}$. The rewriter generates $k$ possible rewritten formats $[\hat{c}_i^{(t)}]_{i=1}^k$ with beam search mechanism. There are two-level external reward functions to evaluate the quality of generated $\hat{c}_i^{(t)}$: token-level reward and sentence-level reward.

\noindent \textbf{Token-level Reward} To reserve the schema information of the database mentioned in original utterance $x^{(t)}$, the generated token $\hat{c}_i^{(t,j)}$ will get $\rm{+0.1}$ reward at $j$-th step when it is database-related token mentioned in $x^{(t)}$. To decrease the co-reference phenomenon in the rewritten utterance, we punish the generated pronoun words (e.g., it, their, and so on) with $\rm{-0.1}$. Otherwise, the generated tokens will get zero points.

\noindent \textbf{Sentence-level Reward} We first use the pretrained language model ${\rm LM}_c(\cdot)$ to evaluate the quality of the rewritten utterance with $r_{\rm LMc}={\rm log}({\rm LM}_c(\hat{c}_i^{(t)})) / {\rm len}(\hat{c}_i^{(t)})$, where ${\rm len}(\hat{c}_i^{(t)})$ denotes the number of the tokens in $\hat{c}_i^{(t)}$. In practice, the rewritten utterance $\hat{c}_i^{(t)}$ can be directly evaluated by the user, who does not need any SQL background. The user can give an indicated score (0 or 1) to evaluate whether $\hat{c}_i^{(t)}$ meets his/her real intent. Instead, we feed the rewritten utterance $\hat{c}_i^{(t)}$ into Text-to-SQL parser and get the corresponding SQL query with $\hat{q} = {\rm RATSQL(\hat{c}_i^{(t)})}$. If $\hat{q}$ equals to the golden SQL, we can say the rewritten utterance meets the user's intent ($r_{\rm u}=1$) and vise versa ($r_{\rm u}=0$). The final sentence-level reward of $\hat{c}_i^{(t)}$ is represented as $r_i^c=r_{\rm LMc}+r_{\rm u}$.

For the $j$-th token in the rewritten utterance $\hat{c}_i^{(t)}$, its accumulated reward can be represented as $R_i^{(t,j)}=r_i^{(t,j)}+\sum_{l=j+1}^{m}\lambda^{l-j}r_i^{(t,l)}$, where $\lambda$ is discount rate, $r_i^{(t,j)}$ means the $j$-th token reward of the rewritten utterance and the final token reward equals to the sentence-level reward $r_i^{(t,m)}=r_i^c$. The rewriter can be optimized by policy gradient method as:
\begin{gather*}
L_{pg}(\Theta_c) = -\sum_{i,j=1}^{k,m}R_i^{(t,j)}{\rm log}\left(P(\hat{c}_i^{(t,j)}|(x^{(t)};h);\Theta_c)\right).
\end{gather*}
To force the simplifier to reconstruct the original input $x^{(t)}$ as similar as possible, the simplifier can be optimized with maximum likelihood estimation (MLE) as:
\begin{gather*}
L_{ce}(\Theta_s) = -{\rm log}\left(P(x^{(t)}|(\hat{c}_i^{(t)};h);\Theta_s)\right).
\end{gather*}
Noting that $x^{(t)}$ could be a semantic-completion utterance. It is not reasonable to force the simplifier to reconstruct a semantic-completion utterance. Thus, we first compare the length of the original utterance ${\rm len}(x^{(t)})$ with the length of the rewritten one ${\rm len}(\hat{c}_i^{(t)})$. Only when ${\rm len}(x^{(t)})<{\rm len}(\hat{c}_i^{(t)})$, we optimize the simplifier with MLE.

\subsubsection{Loop Starts from Simplifier}
As shown in Fig.~\ref{fig:dual_framework}, we also sample an unlabeled dialogue data $(x^{(t)};h)$ from $\mathcal{D}_{u}$. The simplifier generates $k$ possible simplified formats $[\hat{s}_i^{(t)}]_{i=1}^k$ with beam search mechanism evaluated by two-level external reward functions.

\noindent \textbf{Token-level Reward} To decrease the schema information mentioned in original utterance $x^{(t)}$, the generated token $\hat{s}_i^{(t,j)}$ will get $\rm{-0.1}$ punishment at $j$-th step when it is database-related token mentioned in $x^{(t)}$ and history $h$. To encourage the co-reference phenomenon in the simplified utterance, we award the pronoun words with $\rm{+0.1}$ reward. Otherwise, the generated tokens will get zero points.

\noindent \textbf{Sentence-level Reward} We only use the pretrained language model ${\rm LM}_s(\cdot)$ to evaluate the quality of the simplified utterance with $r_{\rm LMs}={\rm log}({\rm LM}_s(\hat{s}_i^{(t)})) / {\rm len}(\hat{s}_i^{(t)})$, where ${\rm len}(\hat{s}_i^{(t)})$ denotes the number of the tokens in $\hat{s}_i^{(t)}$.

For the $j$-th token in the simplified utterance $\hat{s}_i^{(t)}$, its accumulated reward is represented as $R_i^{(t,j)}=r_i^{(t,j)}+\sum_{l=j+1}^{m}\lambda^{l-j}r_i^{(t,l)}$, where $r_i^{(t,j)}$ means the $j$-th token reward of the simplified utterance and the final token reward equals to the sentence-level reward $r_i^{(t,m)}=r_{\rm LMs}$. Similar to the first loop in Section~\ref{sec:loop1}, the simplifier and the rewriter can be optimized with policy gradient method and MLE respectively:
\begin{gather*}
L_{pg}(\Theta_s) = -\sum_{i,j=1}^{k,m}R_i^{(t,j)}{\rm log}\left(P(\hat{s}_i^{(t,j)}|(x^{(t)};h);\Theta_s)\right), \\
L_{ce}(\Theta_c) = -{\rm log}\left(P(x^{(t)}|(\hat{s}_i^{(t)};h);\Theta_c)\right).
\end{gather*}
Noting that only when ${\rm len}(x^{(t)})>{\rm len}(\hat{s}_i^{(t)})$, we optimize the rewriter with MLE.

\section{Experiments}
Series of experiments are conducted to validate our proposed utterance rewrite model and the decoupled framework. We first validate the pretrained BART's performance on the utterance rewrite benchmarks. Then, the multi-turn Text-to-SQL with the decoupled parsing method (DELTA) are experimented on the limited utterance rewrite data. Finally, we analyze the interpretability of the decoupled parsing method through the case study.

\begin{table*}
\begin{minipage}{0.5\linewidth}
\centering
\begin{tabular}{cccc}
  \hline
  \multicolumn{4}{c}{T{\small ASK}} \\
  \hline
  \hline
  \textbf{Models}  & \textbf{EM} & ${\rm \bf{B}}_4$ & $\mathcal{F}_1$ \\
    \hline
    GECOR 1$^\dag$ & 68.5 & 83.9 & 66.1 \\
    GECOR 2$^\dag$ & 66.2 & 83.0 & 66.2 \\
    RUN$^\ddag$ & 69.2 &  85.6 & 70.6 \\
    BART & \textbf{74.2} & \textbf{89.4} & \textbf{81.2} \\
    \hline
  \end{tabular}
\end{minipage}\begin{minipage}{0.5\linewidth}  
\centering
\begin{tabular}{cccccc}
  \hline
  \multicolumn{6}{c}{C{\small ANARD}} \\
  \hline
  \hline
  \textbf{Models}  & ${\rm \bf{B}}_1$ & ${\rm \bf{B}}_2$ & ${\rm \bf{B}}_4$ & ${\rm \bf{R}}_2$ & ${\rm \bf{R}_L}$ \\
    \hline
    Pronoun Sub$^\ddag$ & 60.4 & 55.3 & 47.4 & 63.7 & 73.9 \\
    L-Ptr-Gen$^\ddag$ & 67.2 & 60.3 & 50.2 & 62.9 & 74.9 \\
    RUN$^\ddag$ & 70.5 & 61.2 & 49.1 & 61.2 & 74.7 \\
    BART & \textbf{84.5} & \textbf{71.3} & \textbf{54.3} & \textbf{71.1} & \textbf{81.7} \\
    \hline
  \end{tabular}
\end{minipage}
\caption{The experimental results on rewrite datasets T{\small ASK} (\textbf{left}) and C{\small ANARD} (\textbf{right}). $\dag$: results from ~\cite{quan2019gecor}; $\ddag$: results from ~\cite{liu2020incomplete}.}
\label{tab:rewriter}
\end{table*}

\begin{table*}
\centering
\resizebox{\linewidth}{!}{
\begin{tabular}{ccccccccc}
  \hline
  \multirow{3}*{\textbf{Models}} & \multicolumn{4}{c}{SParC} & \multicolumn{4}{c}{CoSQL} \\
    \cline{2-9}
    ~ & \multicolumn{2}{c}{Question Match} & \multicolumn{2}{c}{Interaction Match} & \multicolumn{2}{c}{Question Match} & \multicolumn{2}{c}{Interaction Match} \\
    ~ & Dev. & Test & Dev. & Test & Dev. & Test & Dev. & Test \\
    \hline
    EditSQL~\cite{zhang2019editing} & 47.2 & 47.9 & 29.5 & 25.3 & 39.9 & 40.8 & 12.3 & 13.7 \\ 
    RichContext~\cite{ijcai2020Liu} & 52.6 & - & 29.9 & - & 41.0 & - & 14.0 & - \\ 
    IGSQL~\cite{caiwan2020igsql} & 50.7 & 51.2 & 32.5 & 29.5 & 44.1 & 42.5 & 15.8 & 15.0 \\
    R$^2$SQL~\cite{hui2021dynamic} & 54.1 & 55.8 & 35.2 & 30.8 & 45.7 & 46.8 & 19.5 & 17.0 \\
    \hline
    DELTA+Dual(ours) & \textbf{58.6} & \textbf{59.9} & \textbf{35.6} & \textbf{31.8} & \textbf{51.7} & \textbf{50.8} & \textbf{21.5} & \textbf{19.7} \\
    \hline
  \end{tabular}}
\caption{The question match accuracy and interaction match accuracy on SParC and CoSQL datasets. Since the test datasets are not public, RichContext has not evaluated by the dataset owner.}
\label{tab:twophasedual}
\end{table*}

\subsection{Experimental Setup}

\noindent \textbf{Datasets\&Metrics} Our proposed rewrite model is validated on two utterance rewrite datasets: T{\small ASK}~\cite{quan2019gecor} and C{\small ANARD}~\cite{elgohary2019can}. We employ the widely used automatic metrics BLEU, ROUGE, EM (Exact Match) and rewrite F-score as our evaluation metrics. ${\rm \bf{BLEU}}_n$(${\rm \bf{B}}_n$) and ${\rm \bf{ROUGE}}_n$(${\rm \bf{R}}_n$) are used to calculate the similarity and the overlapping at the $n$-grams level between predictions and golden ones. \textbf{EM} means the exact match rate, where the prediction exactly equals to the golden. Rewrite F-score $\mathcal{F}_n$ is calculated on the collection of $n$-grams that contain at least one word from the context. Our decoupled parsing method is evaluated on two multi-turn Text-to-SQL tasks: SParC and CoSQL. Following~\cite{Yu&al.19}, with \textbf{Question Match} and \textbf{Interaction Match} as the metrics. Question match means the predicted SQL equals the golden one for each question, while Interaction match indicates the predicted SQL queries of all the questions in an interaction are correct.


\noindent \textbf{Implementation Details} Our implementation is based on PyTorch~\cite{NEURIPS2019_9015} and HuggingFace’s~\cite{wolf2020transformers} Transformers library. We reproduce RATSQL with the same setup presented in ~\cite{rat-sql}, where the encoder consists of eight relation-aware Transformer (RAT) layers. When fine-tuning the BARTs (rewriter and simplifier) on the utterance rewrite datasets, we use the AdamW as the optimizer with the learning rate 2e-6. At the dual learning and co-training period, we set the learning rate as 1e-6. Specifically, BARTs mentioned above refer to ${\rm BART_{large}}$. The discount rate $\lambda$ in the dual learning method is 1.

\subsection{Experimental Results}

\subsubsection{BART as Rewrite Model}
For the rewrite task, we compared the pretrained BART with state-of-the-art rewrite models: L-Ptr-Gen~\cite{see2017get}, GECOR~\cite{quan2019gecor}, and RUN~\cite{liu2020incomplete}.
Table~\ref{tab:rewriter} shows the experimental results on T{\small ASK} and C{\small ANARD} datasets. As indicated, using the BART as rewrite model surpasses the best baseline RUN by a large margin on all the metrics. Even for the most challenging metric \textbf{EM}, the BART exceeds the previous best model by 5.0 points on T{\small ASK}. The BART also obtains a large boost on C{\small ANARD}, which improves the state-of-the-art by 4.1 points and 6.8 points on ${\rm B}_4$ and ${\rm R}_L$ respectively. The above experimental results demonstrate the superiority of the BART as the rewrite model.

\begin{table*}
\centering
\begin{small}
\begin{tabular}{ccccc}
  \hline
  Precedent Question & Current Question & Rewritten Question & Predicted SQL & Status \\
    \hline
    \makecell[l]{how many people \\ live in asia ?} &  \makecell[l]{what about the largest \\ gnp among them ?} & \makecell[l]{\sout{{\color{red} how many people live}} \\ {\color{green} what is the country} \\ in asia that is the \\ largest gnp \sout{{\color{red} among}} \\ \sout{{\color{red} them}} ?} & \makecell[l]{SELECT country.Population \\ FROM country WHERE \\ country.Continent= 'value' \\ ORDER BY country.GNP \\ESC LIMIT 1} & \makecell[c]{{\color{red} Fail} \\ {\color{red} in Phase-\uppercase\expandafter{\romannumeral1}}} \\
    \hline
    \makecell[l]{which students have \\ pets ?} &  \makecell[l]{what are the different \\ first names ?} & \makecell[l]{what are the different \\ first names of the \\students that have pets ?} & \makecell[l]{SELECT Student.Fname \\ FROM Student {\color{green} JOIN} \\ {\color{green} Has\_pet ON Student.stuid} \\ {\color{green} = Has\_pet.stuid}} & \makecell[c]{{\color{red} Fail} \\ {\color{red} in Phase-\uppercase\expandafter{\romannumeral2}}} \\
    \hline
    \makecell[l]{what flights land \\in aberdeen ?} &  \makecell[l]{also include flights that \\ land in abilene .} & \makecell[l]{what flights land in
    \\ aberdeen or abilene ?} & \makecell[l]{SELECT * FROM flights \\ JOIN airports WHERE \\airports.City = 'value' \\OR airports.City = 'value'} & \makecell[c]{{\color{green} Success}} \\
    \hline
  \end{tabular}
\end{small}
\caption{Three instances parsed by our proposed decoupled parsing method with rewritten utterance and final predicted SQL query.
The \sout{{\color{red} red}} means that the error happens. The {\color{green} green} is modified by us.}
\label{tab:example}
\end{table*}

\subsubsection{DELTA for Decoupled Parsing}
Regarding the multi-turn Text-to-SQL task, we compared the decoupled parsing method with all the released end-to-end multi-turn Text-to-SQL models: EditSQL~\cite{zhang2019editing}, RichContext~\cite{ijcai2020Liu}, IGSQL~\cite{caiwan2020igsql}, and R$^2$SQL~\cite{hui2021dynamic}.

Since there is no utterance rewrite data on SParC and CoSQL datasets, we randomly sample 10\% dialogues on these two datasets and annotate them as the rewrite in-domain data. There are 741 annotated turns and 695 annotated turns on SParC and CoSQL respectively. At the Phase-\uppercase\expandafter{\romannumeral1}, we first use the rewrite in-domain data to warm-up the rewrite model and simplification model, where their encoders share the parameters inspired by \cite{lample2018phrase}. Then, we use the rest of 90\% dialogues as unlabeled data to further improve the rewrite model with the dual learning method, detailed in Section~\ref{sec:dual}. At the Phase-\uppercase\expandafter{\romannumeral2}, we first use the single-turn Text-to-SQL data (Spider) to warm-up the RATSQL parser. Since there is an annotation gap\footnote{For example, ``Tell me how many rooms cost more than 120, for each different decor." is annotated as ``SELECT decor, count(*) FROM Rooms WHERE basePrice $>$ 120 GROUP BY decor" in SParC. It is tend to be annotated as ``SELECT count(*) FROM Rooms WHERE basePrice $>$ 120 GROUP BY decor" in Spider.} between Spider and multi-turn Text-to-SQL datasets, we use the annotated Text-to-SQL data on the SParC and CoSQL to fine-tune the pretrained RATSQL parser, where the multi-turn dialogue data are rewritten as single-turn data by the rewrite model trained in Phase-\uppercase\expandafter{\romannumeral1}. Table~\ref{tab:twophasedual} shows that our proposed decoupled framework (DELTA+Dual) gets a considerable performance boost on SParC and CoSQL datasets.

\begin{table}
\centering
\begin{tabular}{lcc}
  \hline
  \multirow{2}*{\textbf{Variants}} & \multicolumn{2}{c}{SParC} \\
    \cline{2-3}
    ~ & QM & IM \\
    \hline
    (0) DELTA + Dual & 58.6 & 35.6 \\
    \hline
    (1) DELTA + Co-training & 57.2 & 33.6 \\ 
    (2) - Dual & 55.5 & 31.7 \\
    (3) - parsing in-domain data & 54.7 & 31.5 \\
    (4) \quad  - rewrite in-domain data & 42.1 & 14.4 \\
    (5) - rewriter & 34.5 & 7.1 \\
    \hline
  \end{tabular}
\caption{The ablations of our proposed decoupled parsing framework. Since the test dataset of the SParC is not released, we report all the performances on its development. QM: question match accuracy; IM: interaction match accuracy.}
\label{tab:ablation}
\vspace{-0.2cm}
\end{table}

\subsubsection{Ablation Study}
We conducted an ablation study to analyze the contribution of our proposed decoupled parsing framework on the SParC dataset. To compare with our proposed dual learning method on rewrite task, we examined another semi-supervised learning method co-training~\cite{blum1998combining}, which uses the pretrained rewrite model to annotate the unlabeled data and add these pseudo-labeled data to improve the original rewrite model iteratively. 
To fairly compare with the dual learning method, we only use the pseudo labeled rewrite data that are correctly predicted by the RATSQL parser at each iteration of the co-training method. As shown in Table~\ref{tab:ablation}, our proposed dual learning method outperforms the co-training method at row (1). To further validate the effect of the dual learning method, we remove the dual learning part in the Phase-\uppercase\expandafter{\romannumeral1}. Compared with our adapted dual learning method at row (2), the above two variants have a significant performance degradation, which demonstrates the superiority of the dual learning method on rewrite task. 

Compared with the end-to-end multi-turn Text-to-SQL models, our proposed decoupled parsing framework even does not require any annotated multi-turn in-domain data. We first evaluate the performance of the decoupled method without any Text-to-SQL parsing in-domain data at row (3). There are 3.9 points and 4.1 points degradation on question match accuracy and interaction match accuracy respectively, which is caused by the annotation gap between Spider and SParC. We further drop our annotated rewrite in-domain data at row (4) and warm-up the rewrite model and simplification model with T{\small ASK} and C{\small ANARD} datasets. As shown in Table~\ref{tab:ablation}, we can find the decoupled parsing framework still gets 42.1\% question match accuracy without any annotated multi-turn in-domain data. Lastly, we just remove the Phase-\uppercase\expandafter{\romannumeral1} (rewrite model) at row (5), where the RATSQL parser is trained on Spider and fine-tuned with multi-turn Text-to-SQL data. It can be regarded as the baseline of all the ablations, which only gets 34.5\% question match accuracy.

\begin{figure}[t]
\centering
\includegraphics[width=0.4\textwidth]{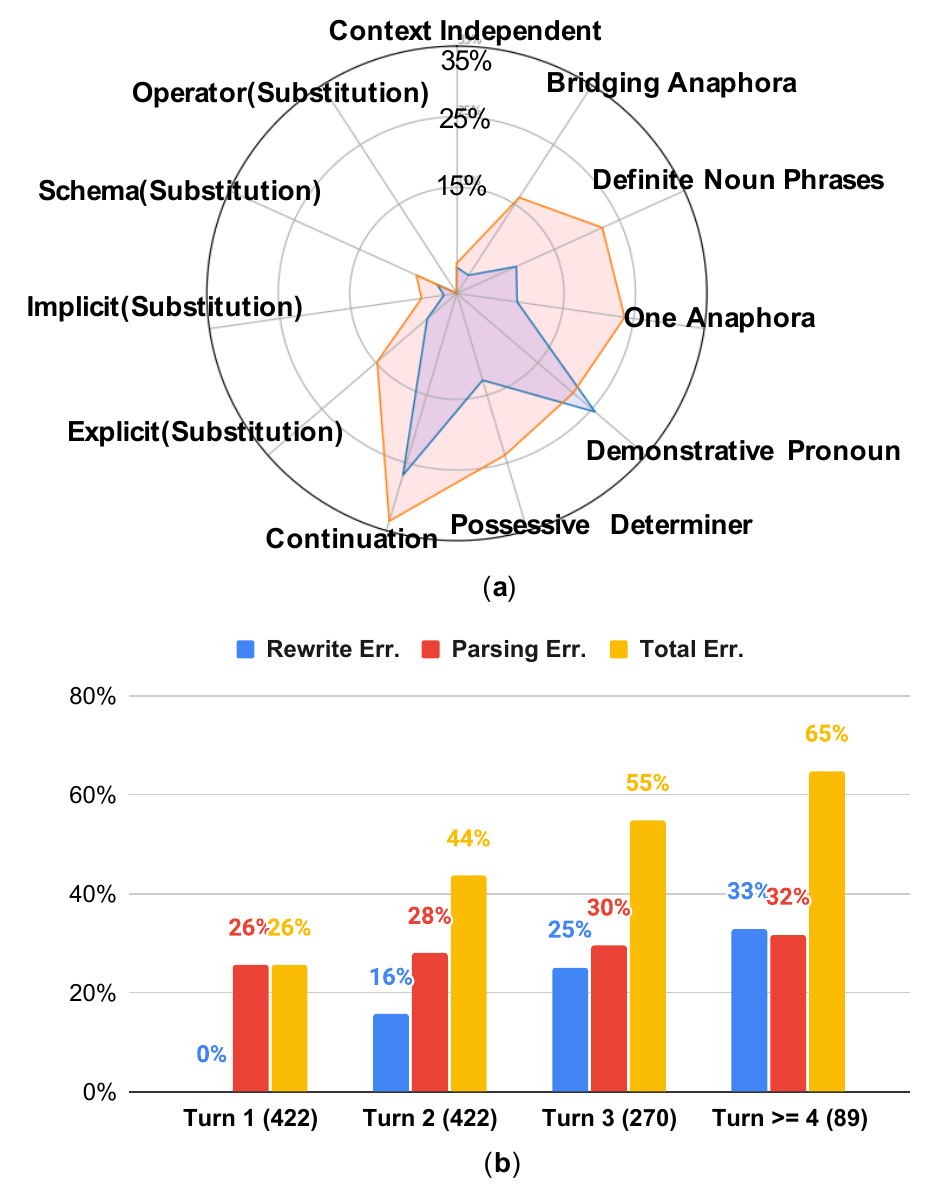} 
\caption{The rewrite error analysis (\textbf{a}) in the fine-grained level and the error rate of the two decoupled models (\textbf{b}) at turn-wise level.}
\label{fig:statis}
\end{figure}

\subsection{Case Analysis}
Compared with end-to-end multi-turn Text-to-SQL models, our decoupled parser can generate the intermediate rewritten utterance, which is easier to understand for the user than a SQL query. As introduced in Section~\ref{sec:loop1}, the feedback of the user can be used to optimize the rewrite model. Additionally, our decoupled parser is more convenient in data collection compared with end-to-end methods, which does not require annotators' familiarization with SQL to rewrite an utterance.
When collecting single-turn Text-to-SQL data, the annotator does not need to consider the dialogue context. It is also costly to collect the dialogue data on the SQL query task. 

Table~\ref{tab:example} displays three cases parsed by our proposed decoupled method. We can pinpoint exactly which phase the error occurred under decoupled parsing framework. Through fine-grained error analysis, the bottleneck of multi-turn parser can be found accurately. Thus, we can target to optimize the bottleneck individually. Figure~\ref{fig:statis}(\textbf{a}) shows the error rate of the utterance rewrite model (DELTA+Dual) on SParC development dataset at a fine-grained level. The orange line denotes the error rate on the individual co-reference or ellipsis type. The blue line denotes the overall error ratio. We can see that most rewrite errors happen on the co-reference side, especially at \textit{Demonstrative Pronoun} type. For the ellipsis, \textit{Continuation} type is a serious problem. Figure~\ref{fig:statis}(\textbf{b}) shows
the error ratios that happen in the rewrite model (Phase-\uppercase\expandafter{\romannumeral1}) or in the parsing model (Phase-\uppercase\expandafter{\romannumeral2}). We can find that at the first three turns the parsing model is still the bottleneck. After the third turn, the rewrite model gets a bigger error rate. The error rate of the rewrite model is more sensitive than the parser with turn increased. We can conclude that we need more annotated rewrite data, especially with \textit{Continuation} type and \textit{Demonstrative Pronoun} type.

\section{Related Work}

\noindent \textbf{Utterance Rewrite} Recently, the utterance rewrite has raised large attention. Some works use the sequence-to-sequence architecture with copy mechanism~\cite{elgohary2019unpack, quan2019gecor, rastogi2019scaling} to solve the incomplete question problem. \citet{liu2019split} decompose the utterance rewrite model as two-phase subtasks: split and recombine. The split and recombine models are both learned from the well-designed reward function by the policy gradient method. Borrowing the idea from image segmentation, \citet{liu2020incomplete} formulate the utterance rewrite as the semantic segmentation task, where the rewrite model is implemented with UNet~\cite{ronneberger2015u}. For the downstream task, the utterance rewrite has been successfully used in dialogue state tracking (DST) tasks~\cite{rastogi2019scaling, han2020multiwoz}. ~\citet{yu2020few} propose a rule-based and self-supervised learning method to generate weakly-supervised rewrite data, which are used to fine-tune GPT-2. Different from the previous works, we directly use the pretrained BART, which is a denoising autoencoder, as the utterance rewrite model.

\noindent \textbf{End-to-End Text-to-SQL Parser} 
EditSQL~\cite{zhang2019editing} proposes an Edit-based model that reuses the SQL query generated from the previous step to alleviate the pressure of the increasing turns. RichContext~\cite{ijcai2020Liu} conducts an exploratory study on semantic parsing
in context and performs a fine-grained analysis. IGSQL~\cite{caiwan2020igsql} presents a schema interaction graph encoder to capture the historical information of database schema items.  R$^2$SQL~\cite{hui2021dynamic} presents a dynamic graph framework that employs dynamic memory decay mechanisms to introduce inductive bias to construct enriched contextual relation representation at both utterance and token level.

\noindent \textbf{Dual Learning} Dual learning method is first proposed to improve neural machine translation (NMT)~\cite{he2016dual}. The dual learning mechanism enables a pair of dual systems to automatically learn from unlabeled data through a closed-loop game. The idea of dual learning has been applied into various tasks, such as Question Answer~\cite{tang2017question}/Generation~\cite{tang2018learning},  Image-to-Image Translation~\cite{yi2017dualgan}, Open-domain Information Extraction/Narration~\cite{sun2018logician}, Text Simplification~\cite{zhao2020semi}, Semantic Parsing~\cite{cao2019semantic,zhu2020dual,cao2020unsupervised} and dialogue state tracking~\cite{chen2020dual}.

\section{Conclusion and Future Work}
In this paper, we propose a decoupled parsing framework (DELTA+Dual) to solve the multi-turn Text-to-SQL task. The previous end-to-end multi-turn Text-to-SQL models rely on large-scale multi-turn data. DELTA can achieve considerable performance without any multi-turn Text-to-SQL data. We adapt the pretrained BART as the rewrite model and achieve new state-of-the-art performance on the utterance rewrite benchmarks. We further propose an efficient dual learning method to make full use of unlabeled dialogue data. On the challenging multi-turn Text-to-SQL benchmarks, DELTA surpasses all the released end-to-end models with fully labeled data. In the future, we will try to reformulate the decoupled parsing method as a multitask, where the rewrite model and Text-to-SQL model are trained simultaneously. The proposed DELTA is also easy to extend to the other conversational semantic parsing tasks, like dialogue state tracking~\cite{chen2020credit,zhu2020efficient,chen2020schema}.

\section*{Acknowledgements}
We thank Bo Pang, Tao Yu and Songhe Wang for their help with the evaluation. We would like to thank the anonymous reviewers for their thoughtful comments. This work has been supported by No. SKLMCPTS2020003 Project, Shanghai Municipal Science and Technology Major Project (2021SHZDZX0102) and Startup Fund for Youngman Research at SJTU (SFYR at SJTU).

\bibliographystyle{acl_natbib}
\bibliography{anthology,acl2021}

\begin{thebibliography}{49}
\expandafter\ifx\csname natexlab\endcsname\relax\def\natexlab#1{#1}\fi

\bibitem[{Andreas et~al.(2020)Andreas, Bufe, Burkett, Chen, Clausman, Crawford,
  Crim, DeLoach, Dorner, Eisner et~al.}]{andreas2020task}
Jacob Andreas, John Bufe, David Burkett, Charles Chen, Josh Clausman, Jean
  Crawford, Kate Crim, Jordan DeLoach, Leah Dorner, Jason Eisner, et~al. 2020.
\newblock Task-oriented dialogue as dataflow synthesis.
\newblock \emph{Transactions of the Association for Computational Linguistics},
  8:556--571.

\bibitem[{Androutsopoulos et~al.(1995)Androutsopoulos, Ritchie, and
  Thanisch}]{androutsopoulos1995natural}
Ion Androutsopoulos, Graeme~D Ritchie, and Peter Thanisch. 1995.
\newblock Natural language interfaces to databases-an introduction.
\newblock \emph{arXiv preprint cmp-lg/9503016}.

\bibitem[{Ba et~al.(2016)Ba, Kiros, and Hinton}]{ba2016layer}
Jimmy~Lei Ba, Jamie~Ryan Kiros, and Geoffrey~E Hinton. 2016.
\newblock Layer normalization.
\newblock \emph{arXiv preprint arXiv:1607.06450}.

\bibitem[{Blum and Mitchell(1998)}]{blum1998combining}
Avrim Blum and Tom Mitchell. 1998.
\newblock Combining labeled and unlabeled data with co-training.
\newblock In \emph{Proceedings of the eleventh annual conference on
  Computational learning theory}, pages 92--100.

\bibitem[{Cai and Wan(2020)}]{caiwan2020igsql}
Yitao Cai and Xiaojun Wan. 2020.
\newblock \href {https://doi.org/10.18653/v1/2020.emnlp-main.560} {{IGSQL}:
  Database schema interaction graph based neural model for context-dependent
  text-to-{SQL} generation}.
\newblock In \emph{Proceedings of the 2020 Conference on Empirical Methods in
  Natural Language Processing (EMNLP)}, pages 6903--6912, Online. Association
  for Computational Linguistics.

\bibitem[{Cao et~al.(2019)Cao, Zhu, Liu, Li, and Yu}]{cao2019semantic}
Ruisheng Cao, Su~Zhu, Chen Liu, Jieyu Li, and Kai Yu. 2019.
\newblock Semantic parsing with dual learning.
\newblock In \emph{Proceedings of the 57th Annual Meeting of the Association
  for Computational Linguistics}, pages 51--64.

\bibitem[{Cao et~al.(2020)Cao, Zhu, Yang, Liu, Ma, Zhao, Chen, and
  Yu}]{cao2020unsupervised}
Ruisheng Cao, Su~Zhu, Chenyu Yang, Chen Liu, Rao Ma, Yanbin Zhao, Lu~Chen, and
  Kai Yu. 2020.
\newblock Unsupervised dual paraphrasing for two-stage semantic parsing.
\newblock In \emph{Proceedings of the 58th Annual Meeting of the Association
  for Computational Linguistics}, pages 6806--6817.

\bibitem[{Chen et~al.(2020{\natexlab{a}})Chen, Lv, Wang, Zhu, Tan, and
  Yu}]{chen2020schema}
Lu~Chen, Boer Lv, Chi Wang, Su~Zhu, Bowen Tan, and Kai Yu. 2020{\natexlab{a}}.
\newblock Schema-guided multi-domain dialogue state tracking with graph
  attention neural networks.
\newblock In \emph{Proceedings of the AAAI Conference on Artificial
  Intelligence}, volume~34, pages 7521--7528.

\bibitem[{Chen et~al.(2020{\natexlab{b}})Chen, Chen, Xu, Zhao, Zhu, and
  Yu}]{chen2020credit}
Zhi Chen, Lu~Chen, Zihan Xu, Yanbin Zhao, Su~Zhu, and Kai Yu.
  2020{\natexlab{b}}.
\newblock Credit: Coarse-to-fine sequence generation for dialogue state
  tracking.
\newblock \emph{arXiv preprint arXiv:2009.10435}.

\bibitem[{Chen et~al.(2021)Chen, Chen, Zhao, Cao, Xu, Zhu, and
  Yu}]{chen-etal-2021-shadowgnn}
Zhi Chen, Lu~Chen, Yanbin Zhao, Ruisheng Cao, Zihan Xu, Su~Zhu, and Kai Yu.
  2021.
\newblock \href {https://www.aclweb.org/anthology/2021.naacl-main.441}
  {{S}hadow{GNN}: Graph projection neural network for text-to-{SQL} parser}.
\newblock In \emph{Proceedings of the 2021 Conference of the North American
  Chapter of the Association for Computational Linguistics: Human Language
  Technologies}, pages 5567--5577, Online. Association for Computational
  Linguistics.

\bibitem[{Chen et~al.(2020{\natexlab{c}})Chen, Chen, Zhao, Zhu, and
  Yu}]{chen2020dual}
Zhi Chen, Lu~Chen, Yanbin Zhao, Su~Zhu, and Kai Yu. 2020{\natexlab{c}}.
\newblock Dual learning for dialogue state tracking.
\newblock \emph{arXiv preprint arXiv:2009.10430}.

\bibitem[{Dahl et~al.(1994)Dahl, Bates, Brown, Fisher, Hunicke-Smith, Pallett,
  Pao, Rudnicky, and Shriberg}]{dahl1994expanding}
Deborah~A Dahl, Madeleine Bates, Michael~K Brown, William~M Fisher, Kate
  Hunicke-Smith, David~S Pallett, Christine Pao, Alexander Rudnicky, and
  Elizabeth Shriberg. 1994.
\newblock Expanding the scope of the atis task: The atis-3 corpus.
\newblock In \emph{HUMAN LANGUAGE TECHNOLOGY: Proceedings of a Workshop held at
  Plainsboro, New Jersey, March 8-11, 1994}.

\bibitem[{Elgohary et~al.(2019{\natexlab{a}})Elgohary, Peskov, and
  Boyd-Graber}]{elgohary2019can}
Ahmed Elgohary, Denis Peskov, and Jordan Boyd-Graber. 2019{\natexlab{a}}.
\newblock Can you unpack that? learning to rewrite questions-in-context.
\newblock In \emph{Proceedings of the 2019 Conference on Empirical Methods in
  Natural Language Processing and the 9th International Joint Conference on
  Natural Language Processing (EMNLP-IJCNLP)}, pages 5920--5926.

\bibitem[{Elgohary et~al.(2019{\natexlab{b}})Elgohary, Peskov, and
  Boyd-Graber}]{elgohary2019unpack}
Ahmed Elgohary, Denis Peskov, and Jordan Boyd-Graber. 2019{\natexlab{b}}.
\newblock \href {https://doi.org/10.18653/v1/D19-1605} {Can you unpack that?
  learning to rewrite questions-in-context}.
\newblock In \emph{Proceedings of the 2019 Conference on Empirical Methods in
  Natural Language Processing and the 9th International Joint Conference on
  Natural Language Processing (EMNLP-IJCNLP)}, pages 5918--5924, Hong Kong,
  China. Association for Computational Linguistics.

\bibitem[{Han et~al.(2020)Han, Liu, Takanobu, Lian, Huang, Peng, and
  Huang}]{han2020multiwoz}
Ting Han, Ximing Liu, Ryuichi Takanobu, Yixin Lian, Chongxuan Huang, Wei Peng,
  and Minlie Huang. 2020.
\newblock Multiwoz-coref: A multi-domain task-oriented dataset enhanced with
  annotation corrections and co-reference annotation.
\newblock \emph{arXiv preprint arXiv:2010.05594}.

\bibitem[{He et~al.(2016)He, Xia, Qin, Wang, Yu, Liu, and Ma}]{he2016dual}
Di~He, Yingce Xia, Tao Qin, Liwei Wang, Nenghai Yu, Tie-Yan Liu, and Wei-Ying
  Ma. 2016.
\newblock Dual learning for machine translation.
\newblock In \emph{Advances in neural information processing systems}, pages
  820--828.

\bibitem[{Hui et~al.(2021)Hui, Geng, Ren, Li, Li, Sun, Huang, Si, Zhu, and
  Zhu}]{hui2021dynamic}
Binyuan Hui, Ruiying Geng, Qiyu Ren, Binhua Li, Yongbin Li, Jian Sun, Fei
  Huang, Luo Si, Pengfei Zhu, and Xiaodan Zhu. 2021.
\newblock Dynamic hybrid relation network for cross-domain context-dependent
  semantic parsing.
\newblock \emph{arXiv preprint arXiv:2101.01686}.

\bibitem[{Iyyer et~al.(2017)Iyyer, Yih, and Chang}]{iyyer2017search}
Mohit Iyyer, Wen-tau Yih, and Ming-Wei Chang. 2017.
\newblock Search-based neural structured learning for sequential question
  answering.
\newblock In \emph{Proceedings of the 55th Annual Meeting of the Association
  for Computational Linguistics (Volume 1: Long Papers)}, pages 1821--1831.

\bibitem[{Lample et~al.(2018)Lample, Ott, Conneau, Denoyer, and
  Ranzato}]{lample2018phrase}
Guillaume Lample, Myle Ott, Alexis Conneau, Ludovic Denoyer, and Marc’Aurelio
  Ranzato. 2018.
\newblock Phrase-based \& neural unsupervised machine translation.
\newblock In \emph{Proceedings of the 2018 Conference on Empirical Methods in
  Natural Language Processing}, pages 5039--5049.

\bibitem[{Lewis et~al.(2020)Lewis, Liu, Goyal, Ghazvininejad, Mohamed, Levy,
  Stoyanov, and Zettlemoyer}]{lewis2020bart}
Mike Lewis, Yinhan Liu, Naman Goyal, Marjan Ghazvininejad, Abdelrahman Mohamed,
  Omer Levy, Veselin Stoyanov, and Luke Zettlemoyer. 2020.
\newblock Bart: Denoising sequence-to-sequence pre-training for natural
  language generation, translation, and comprehension.
\newblock In \emph{Proceedings of the 58th Annual Meeting of the Association
  for Computational Linguistics}, pages 7871--7880.

\bibitem[{Liu et~al.(2020{\natexlab{a}})Liu, Chen, Guo, Lou, Zhou, and
  Zhang}]{ijcai2020Liu}
Qian Liu, Bei Chen, Jiaqi Guo, Jian-Guang Lou, Bin Zhou, and Dongmei Zhang.
  2020{\natexlab{a}}.
\newblock \href {https://doi.org/10.24963/ijcai.2020/495} {How far are we from
  effective context modeling? an exploratory study on semantic parsing in
  context}.
\newblock In \emph{Proceedings of the Twenty-Ninth International Joint
  Conference on Artificial Intelligence, {IJCAI-20}}, pages 3580--3586.
  International Joint Conferences on Artificial Intelligence Organization.

\bibitem[{Liu et~al.(2019)Liu, Chen, Liu, Lou, Fang, Zhou, and
  Zhang}]{liu2019split}
Qian Liu, Bei Chen, Haoyan Liu, Jian-Guang Lou, Lei Fang, Bin Zhou, and Dongmei
  Zhang. 2019.
\newblock \href {https://doi.org/10.18653/v1/D19-1535} {A split-and-recombine
  approach for follow-up query analysis}.
\newblock In \emph{Proceedings of the 2019 Conference on Empirical Methods in
  Natural Language Processing and the 9th International Joint Conference on
  Natural Language Processing (EMNLP-IJCNLP)}, pages 5316--5326, Hong Kong,
  China. Association for Computational Linguistics.

\bibitem[{Liu et~al.(2020{\natexlab{b}})Liu, Chen, Lou, Zhou, and
  Zhang}]{liu2020incomplete}
Qian Liu, Bei Chen, Jian-Guang Lou, Bin Zhou, and Dongmei Zhang.
  2020{\natexlab{b}}.
\newblock Incomplete utterance rewriting as semantic segmentation.
\newblock In \emph{Proceedings of the 2020 Conference on Empirical Methods in
  Natural Language Processing (EMNLP)}, pages 2846--2857.

\bibitem[{Paszke et~al.(2019)Paszke, Gross, Massa, Lerer, Bradbury, Chanan,
  Killeen, Lin, Gimelshein, Antiga, Desmaison, Kopf, Yang, DeVito, Raison,
  Tejani, Chilamkurthy, Steiner, Fang, Bai, and Chintala}]{NEURIPS2019_9015}
Adam Paszke, Sam Gross, Francisco Massa, Adam Lerer, James Bradbury, Gregory
  Chanan, Trevor Killeen, Zeming Lin, Natalia Gimelshein, Luca Antiga, Alban
  Desmaison, Andreas Kopf, Edward Yang, Zachary DeVito, Martin Raison, Alykhan
  Tejani, Sasank Chilamkurthy, Benoit Steiner, Lu~Fang, Junjie Bai, and Soumith
  Chintala. 2019.
\newblock \href
  {http://papers.neurips.cc/paper/9015-pytorch-an-imperative-style-high-performance-deep-learning-library.pdf}
  {Pytorch: An imperative style, high-performance deep learning library}.
\newblock In H.~Wallach, H.~Larochelle, A.~Beygelzimer, F.~d\textquotesingle
  Alch\'{e}-Buc, E.~Fox, and R.~Garnett, editors, \emph{Advances in Neural
  Information Processing Systems 32}, pages 8024--8035. Curran Associates, Inc.

\bibitem[{Quan et~al.(2019)Quan, Xiong, Webber, and Hu}]{quan2019gecor}
Jun Quan, Deyi Xiong, Bonnie Webber, and Changjian Hu. 2019.
\newblock Gecor: An end-to-end generative ellipsis and co-reference resolution
  model for task-oriented dialogue.
\newblock In \emph{Proceedings of the 2019 Conference on Empirical Methods in
  Natural Language Processing and the 9th International Joint Conference on
  Natural Language Processing (EMNLP-IJCNLP)}, pages 4539--4549.

\bibitem[{Radford et~al.(2019)Radford, Wu, Child, Luan, Amodei, and
  Sutskever}]{radford2019language}
Alec Radford, Jeff Wu, Rewon Child, David Luan, Dario Amodei, and Ilya
  Sutskever. 2019.
\newblock Language models are unsupervised multitask learners.

\bibitem[{Rastogi et~al.(2019)Rastogi, Gupta, Chen, and
  Mathias}]{rastogi2019scaling}
Pushpendre Rastogi, Arpit Gupta, Tongfei Chen, and Lambert Mathias. 2019.
\newblock Scaling multi-domain dialogue state tracking via query reformulation.
\newblock In \emph{Proceedings of the 2019 Conference of the North American
  Chapter of the Association for Computational Linguistics}. Association for
  Computational Linguistics.

\bibitem[{Ronneberger et~al.(2015)Ronneberger, Fischer, and
  Brox}]{ronneberger2015u}
Olaf Ronneberger, Philipp Fischer, and Thomas Brox. 2015.
\newblock U-net: Convolutional networks for biomedical image segmentation.
\newblock In \emph{International Conference on Medical image computing and
  computer-assisted intervention}, pages 234--241. Springer.

\bibitem[{Rubin and Berant(2021)}]{rubin-berant-2021-smbop}
Ohad Rubin and Jonathan Berant. 2021.
\newblock \href {https://www.aclweb.org/anthology/2021.naacl-main.29}
  {{S}m{B}o{P}: Semi-autoregressive bottom-up semantic parsing}.
\newblock In \emph{Proceedings of the 2021 Conference of the North American
  Chapter of the Association for Computational Linguistics: Human Language
  Technologies}, pages 311--324, Online. Association for Computational
  Linguistics.

\bibitem[{See et~al.(2017)See, Liu, and Manning}]{see2017get}
Abigail See, Peter~J Liu, and Christopher~D Manning. 2017.
\newblock Get to the point: Summarization with pointer-generator networks.
\newblock In \emph{Proceedings of the 55th Annual Meeting of the Association
  for Computational Linguistics (Volume 1: Long Papers)}, pages 1073--1083.

\bibitem[{Shi et~al.(2020)Shi, Ng, Wang, Zhu, Li, Wang, Santos, and
  Xiang}]{shi2020learning}
Peng Shi, Patrick Ng, Zhiguo Wang, Henghui Zhu, Alexander~Hanbo Li, Jun Wang,
  Cicero Nogueira~dos Santos, and Bing Xiang. 2020.
\newblock Learning contextual representations for semantic parsing with
  generation-augmented pre-training.
\newblock \emph{arXiv preprint arXiv:2012.10309}.

\bibitem[{Sun et~al.(2018)Sun, Li, and Li}]{sun2018logician}
Mingming Sun, Xu~Li, and Ping Li. 2018.
\newblock Logician and orator: Learning from the duality between language and
  knowledge in open domain.
\newblock In \emph{Proceedings of the 2018 Conference on Empirical Methods in
  Natural Language Processing}, pages 2119--2130.

\bibitem[{Sutton et~al.(1999)Sutton, McAllester, Singh, Mansour
  et~al.}]{sutton1999policy}
Richard~S Sutton, David~A McAllester, Satinder~P Singh, Yishay Mansour, et~al.
  1999.
\newblock Policy gradient methods for reinforcement learning with function
  approximation.
\newblock In \emph{NIPs}, volume~99, pages 1057--1063. Citeseer.

\bibitem[{Tang et~al.(2017)Tang, Duan, Qin, Yan, and Zhou}]{tang2017question}
Duyu Tang, Nan Duan, Tao Qin, Zhao Yan, and Ming Zhou. 2017.
\newblock Question answering and question generation as dual tasks.
\newblock \emph{arXiv preprint arXiv:1706.02027}.

\bibitem[{Tang et~al.(2018)Tang, Duan, Yan, Zhang, Sun, Liu, Lv, and
  Zhou}]{tang2018learning}
Duyu Tang, Nan Duan, Zhao Yan, Zhirui Zhang, Yibo Sun, Shujie Liu, Yuanhua Lv,
  and Ming Zhou. 2018.
\newblock Learning to collaborate for question answering and asking.
\newblock In \emph{Proceedings of the 2018 Conference of the North American
  Chapter of the Association for Computational Linguistics: Human Language
  Technologies, Volume 1 (Long Papers)}, pages 1564--1574.

\bibitem[{Vaswani et~al.(2017)Vaswani, Shazeer, Parmar, Uszkoreit, Jones,
  Gomez, Kaiser, and Polosukhin}]{vaswani2017attention}
Ashish Vaswani, Noam Shazeer, Niki Parmar, Jakob Uszkoreit, Llion Jones,
  Aidan~N Gomez, {\L}ukasz Kaiser, and Illia Polosukhin. 2017.
\newblock Attention is all you need.
\newblock In \emph{Advances in neural information processing systems}, pages
  5998--6008.

\bibitem[{Wang et~al.(2020)Wang, Shin, Liu, Polozov, and Richardson}]{rat-sql}
Bailin Wang, Richard Shin, Xiaodong Liu, Oleksandr Polozov, and Matthew
  Richardson. 2020.
\newblock {RAT-SQL}: Relation-aware schema encoding and linking for
  text-to-{SQL} parsers.
\newblock In \emph{Proceedings of the 58th Annual Meeting of the Association
  for Computational Linguistics}, pages 7567--7578, Online. Association for
  Computational Linguistics.

\bibitem[{Wolf et~al.(2020)Wolf, Debut, Sanh, Chaumond, Delangue, Moi, Cistac,
  Rault, Louf, Funtowicz, Davison, Shleifer, von Platen, Ma, Jernite, Plu, Xu,
  Scao, Gugger, Drame, Lhoest, and Rush}]{wolf2020transformers}
Thomas Wolf, Lysandre Debut, Victor Sanh, Julien Chaumond, Clement Delangue,
  Anthony Moi, Pierric Cistac, Tim Rault, Rémi Louf, Morgan Funtowicz, Joe
  Davison, Sam Shleifer, Patrick von Platen, Clara Ma, Yacine Jernite, Julien
  Plu, Canwen Xu, Teven~Le Scao, Sylvain Gugger, Mariama Drame, Quentin Lhoest,
  and Alexander~M. Rush. 2020.
\newblock \href {https://www.aclweb.org/anthology/2020.emnlp-demos.6}
  {Transformers: State-of-the-art natural language processing}.
\newblock In \emph{Proceedings of the 2020 Conference on Empirical Methods in
  Natural Language Processing: System Demonstrations}, pages 38--45, Online.
  Association for Computational Linguistics.

\bibitem[{Yi et~al.(2017)Yi, Zhang, Tan, and Gong}]{yi2017dualgan}
Zili Yi, Hao Zhang, Ping Tan, and Minglun Gong. 2017.
\newblock Dualgan: Unsupervised dual learning for image-to-image translation.
\newblock In \emph{Proceedings of the IEEE international conference on computer
  vision}, pages 2849--2857.

\bibitem[{Yu et~al.(2020{\natexlab{a}})Yu, Liu, Yang, Xiong, Bennett, Gao, and
  Liu}]{yu2020few}
Shi Yu, Jiahua Liu, Jingqin Yang, Chenyan Xiong, Paul Bennett, Jianfeng Gao,
  and Zhiyuan Liu. 2020{\natexlab{a}}.
\newblock Few-shot generative conversational query rewriting.
\newblock In \emph{Proceedings of the 43rd International ACM SIGIR Conference
  on Research and Development in Information Retrieval}, pages 1933--1936.

\bibitem[{Yu et~al.(2020{\natexlab{b}})Yu, Wu, Lin, Wang, Tan, Yang, Radev,
  Socher, and Xiong}]{yu2020grappa}
Tao Yu, Chien-Sheng Wu, Xi~Victoria Lin, Bailin Wang, Yi~Chern Tan, Xinyi Yang,
  Dragomir Radev, Richard Socher, and Caiming Xiong. 2020{\natexlab{b}}.
\newblock Grappa: Grammar-augmented pre-training for table semantic parsing.
\newblock \emph{arXiv preprint arXiv:2009.13845}.

\bibitem[{Yu et~al.(2019{\natexlab{a}})Yu, Zhang, Er, Li, Xue, Pang, Lin, Tan,
  Shi, Li et~al.}]{yu2019cosql}
Tao Yu, Rui Zhang, Heyang Er, Suyi Li, Eric Xue, Bo~Pang, Xi~Victoria Lin,
  Yi~Chern Tan, Tianze Shi, Zihan Li, et~al. 2019{\natexlab{a}}.
\newblock Cosql: A conversational text-to-sql challenge towards cross-domain
  natural language interfaces to databases.
\newblock In \emph{Proceedings of the 2019 Conference on Empirical Methods in
  Natural Language Processing and the 9th International Joint Conference on
  Natural Language Processing (EMNLP-IJCNLP)}, pages 1962--1979.

\bibitem[{Yu et~al.(2018)Yu, Zhang, Yang, Yasunaga, Wang, Li, Ma, Li, Yao,
  Roman, Zhang, and Radev}]{yu2018spider}
Tao Yu, Rui Zhang, Kai Yang, Michihiro Yasunaga, Dongxu Wang, Zifan Li, James
  Ma, Irene Li, Qingning Yao, Shanelle Roman, Zilin Zhang, and Dragomir Radev.
  2018.
\newblock Spider: A large-scale human-labeled dataset for complex and
  cross-domain semantic parsing and text-to-sql task.
\newblock In \emph{Proceedings of the 2018 Conference on Empirical Methods in
  Natural Language Processing}, Brussels, Belgium. Association for
  Computational Linguistics.

\bibitem[{Yu et~al.(2019{\natexlab{b}})Yu, Zhang, Yasunaga, Tan, Lin, Li,
  Heyang~Er, Pang, Chen, Ji, Dixit, Proctor, Shim, Jonathan~Kraft, Xiong,
  Socher, and Radev}]{Yu&al.19}
Tao Yu, Rui Zhang, Michihiro Yasunaga, Yi~Chern Tan, Xi~Victoria Lin, Suyi Li,
  Irene~Li Heyang~Er, Bo~Pang, Tao Chen, Emily Ji, Shreya Dixit, David Proctor,
  Sungrok Shim, Vincent~Zhang Jonathan~Kraft, Caiming Xiong, Richard Socher,
  and Dragomir Radev. 2019{\natexlab{b}}.
\newblock Sparc: Cross-domain semantic parsing in context.
\newblock In \emph{Proceedings of the 57th Annual Meeting of the Association
  for Computational Linguistics}, Florence, Italy. Association for
  Computational Linguistics.

\bibitem[{Zhang et~al.(2019)Zhang, Yu, Er, Shim, Xue, Lin, Shi, Xiong, Socher,
  and Radev}]{zhang2019editing}
Rui Zhang, Tao Yu, Heyang Er, Sungrok Shim, Eric Xue, Xi~Victoria Lin, Tianze
  Shi, Caiming Xiong, Richard Socher, and Dragomir Radev. 2019.
\newblock \href {https://doi.org/10.18653/v1/D19-1537} {Editing-based {SQL}
  query generation for cross-domain context-dependent questions}.
\newblock In \emph{Proceedings of the 2019 Conference on Empirical Methods in
  Natural Language Processing and the 9th International Joint Conference on
  Natural Language Processing (EMNLP-IJCNLP)}, pages 5338--5349, Hong Kong,
  China. Association for Computational Linguistics.

\bibitem[{Zhao et~al.(2020)Zhao, Chen, Chen, and Yu}]{zhao2020semi}
Yanbin Zhao, Lu~Chen, Zhi Chen, and Kai Yu. 2020.
\newblock Semi-supervised text simplification with back-translation and
  asymmetric denoising autoencoders.
\newblock In \emph{Proceedings of the AAAI Conference on Artificial
  Intelligence}, volume~34, pages 9668--9675.

\bibitem[{Zhong et~al.(2017)Zhong, Xiong, and Socher}]{zhongSeq2SQL2017}
Victor Zhong, Caiming Xiong, and Richard Socher. 2017.
\newblock Seq2sql: Generating structured queries from natural language using
  reinforcement learning.
\newblock \emph{CoRR}, abs/1709.00103.

\bibitem[{Zhu et~al.(2020{\natexlab{a}})Zhu, Cao, and Yu}]{zhu2020dual}
Su~Zhu, Ruisheng Cao, and Kai Yu. 2020{\natexlab{a}}.
\newblock Dual learning for semi-supervised natural language understanding.
\newblock \emph{IEEE/ACM Transactions on Audio, Speech, and Language
  Processing}, 28:1936--1947.

\bibitem[{Zhu et~al.(2020{\natexlab{b}})Zhu, Li, Chen, and
  Yu}]{zhu2020efficient}
Su~Zhu, Jieyu Li, Lu~Chen, and Kai Yu. 2020{\natexlab{b}}.
\newblock Efficient context and schema fusion networks for multi-domain
  dialogue state tracking.
\newblock In \emph{Proceedings of the 2020 Conference on Empirical Methods in
  Natural Language Processing: Findings}, pages 766--781.

\end{thebibliography}


\end{document}